\documentclass[10pt,twocolumn,letterpaper]{article}
\usepackage[accsupp]{axessibility}
\usepackage{float}
\usepackage{balance}
\usepackage{colortbl}
\usepackage{pifont}
\usepackage{amsmath}

\newcommand{\cmark}{\ding{51}}%
\newcommand{\xmark}{\ding{55}}%
\usepackage{gradient-text}
\usepackage[pdftex]{graphicx}
\usepackage{epstopdf}

\definecolor{VeryLightPink}{RGB}{255, 248, 248}
\definecolor{VeryLightBlue}{RGB}{248, 248, 255}
\definecolor{VeryLightGreen}{RGB}{248, 255, 248}

\usepackage{cvpr}              

\usepackage{colortbl}
\usepackage{lipsum}
\usepackage{array}
\usepackage[dvipsnames]{xcolor}

\usepackage{multirow}
\usepackage{appendix}

\definecolor{cvprblue}{rgb}{0.21,0.49,0.74}
\definecolor{firstcolor}{rgb}{0.75686275,0.26666667,0.05490196}
\definecolor{secondcolor}{rgb}{0.29411765,0.0,0.50980392}
\definecolor{mygray}{gray}{0.9}
\definecolor{urlcolor1}{rgb}{0.2039, 0.5569, 0.2431}
\definecolor{urlcolor2}{rgb}{0.2941, 0.0000, 0.5098}
\definecolor{urlcolor3}{rgb}{0.7569, 0.2667, 0.0549}

\usepackage{bbding} 
\usepackage[pagebackref,breaklinks,colorlinks,citecolor=cvprblue]{hyperref}

\def\paperID{4609} 
\def\confName{CVPR}
\def\confYear{2024}

\hypersetup{urlcolor=urlcolor1}

\title{Multiagent Multitraversal Multimodal Self-Driving: \\ \gradientRGB{Open MARS Dataset}{75,0,130}{193,68,14}}

\author{Yiming Li \quad Zhiheng Li \quad Nuo Chen \quad Moonjun Gong \\ Zonglin Lyu  \quad Zehong Wang \quad Peili Jiang \quad Chen Feng\textsuperscript{\ding{41}}\\
New York University\\
{\tt\small yimingli@nyu.edu, cfeng@nyu.edu} \\
}

\begin{document}
\twocolumn[{%
\maketitle
\vspace{-8mm}
\begin{figure}[H]
\begin{center}
\hsize=\textwidth 
\includegraphics[width=\textwidth]{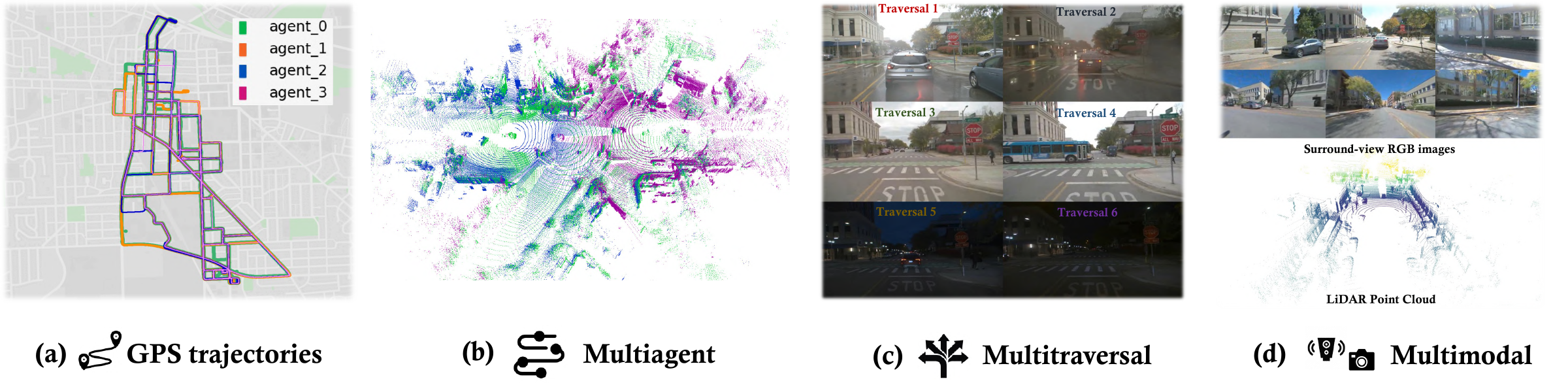}
\vspace{-1mm}
\caption{\textbf{Overview of MARS.} \textbf{(a)} Within a geographical area, we operate four autonomous vehicles, displaying their GPS trajectories from a single day using different colors. \textbf{(b)} Vehicles occasionally come close together (visualized via distinct-colored point clouds), supporting research in multiagent systems. \textbf{(c)} We collect sensory data from repeated traversals of the same location under varying conditions, for learning and perception research with retrospective memory. \textbf{(d)} The dataset includes surround-view RGB images and LiDAR point clouds for cross-modal perception and learning. Note that our data is obtained from May Mobility~\includegraphics[height=1.1em]{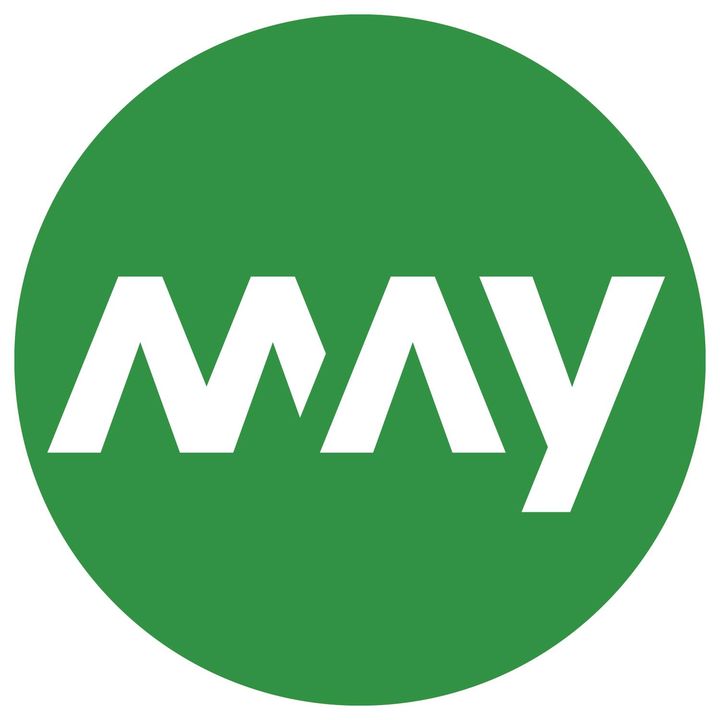}~: \url{https://maymobility.com/}}.
\label{fig:teaser}
\end{center} 
\vspace{-5mm}
\end{figure}
}]
\hypersetup{urlcolor=cvprblue}
\begin{abstract}
\vspace{-2mm}
Large-scale datasets have fueled recent advancements in AI-based autonomous vehicle research. However, these datasets are usually collected from a single vehicle's one-time pass of a certain location, lacking multiagent interactions or repeated traversals of the same place. Such information could lead to transformative enhancements in autonomous vehicles' perception, prediction, and planning capabilities. To bridge this gap, in collaboration with the self-driving company \emph{May Mobility}, we present the \textbf{MARS} dataset which unifies scenarios that enable \textbf{M}ulti\textbf{A}gent, multitrave\textbf{RS}al, and multimodal autonomous vehicle research. More specifically, MARS is collected with a fleet of autonomous vehicles driving within a certain geographical area. Each vehicle has its own route and different vehicles may appear at nearby locations. Each vehicle is equipped with a LiDAR and surround-view RGB cameras. We curate two subsets in MARS: one facilitates collaborative driving with multiple vehicles simultaneously present at the same location, and the other enables memory retrospection through asynchronous traversals of the same location by multiple vehicles. We conduct experiments in place recognition and neural reconstruction. More importantly, MARS introduces new research opportunities and challenges such as multitraversal 3D reconstruction, multiagent perception, and unsupervised object discovery. Our data and codes can be found at \url{https://ai4ce.github.io/MARS/}.
\vspace{-4mm}
\end{abstract}

\begin{table*}[t]
\scriptsize
    \centering
    \renewcommand\tabcolsep{11pt}
    \caption{\textbf{Comparison of existing autonomous driving datasets with multimodal sensors.} C denotes the camera and L denotes LiDAR.}
    \label{tab:datasets}
    \begin{tabular}{cccccccc}
        \toprule
        \textbf{Datasets}  & \textbf{Sensors}   & \textbf{Camera view} & \textbf{Location} & \textbf{Source} & \textbf{Year} & \textbf{Multiagent} &  \textbf{Multitraversal} \\\midrule
        
         KITTI~\cite{geiger2012we}  & C\&L   & Front & Germany & Academia &  2012 & \xmark & \xmark \\
         Lyft Level 5~\cite{WovenPlanetHoldingsInc2019}  & C\&L  & Surround & U.S. & Industry & 2019 &  \xmark & \cmark \\ 
         Argoverse~\cite{chang2019argoverse,wilson2021argoverse}  & C\&L  & Surround & U.S. & Industry &  2019\&2021 &  \xmark & \cmark \\ 
         ApolloScape~\cite{huang2019apolloscape}  & C\&L  & Front & China & Industry &  2019 & \xmark  & \xmark \\ 
         A2D2~\cite{geyer2020a2d2}  & C\&L  & Surround & Germany & Industry &   2020 & \xmark & \xmark \\ 
         A*3D~\cite{pham20203d}  & C\&L  & Front & SG & Academia &  2020 &  \xmark & \xmark \\ 
          {nuScenes}~\cite{caesar2020nuscenes}  & C\&L   & Surround & U.S. \& SG & Industry &  2020 &  \xmark & \cmark \\
          {Waymo Open Dataset}~\cite{sun2020scalability} & C\&L  & Surround & U.S. & Industry &  2020 & \xmark & \xmark \\
         ONCE~\cite{mao1one}  & C\&L  & Surround & China & Industry &  2021 &  \xmark & \xmark\\ 
          {KITTI-360}~\cite{liao2022kitti} & C\&L  & Surround & Germany & Academia &  2022 & \xmark & \xmark  \\ 
         Ithaca365~\cite{diaz2022ithaca365}  & C\&L    & Front & U.S. & Academia &  2022 & \xmark & \cmark \\
         V2V4Real~\cite{xu2023v2v4real}  & C\&L    & Front\&Back & U.S. & Academia &  2023 & \cmark & \xmark \\
          Zenseact Open Dataset~\cite{alibeigi2023zenseact}  & C\&L   & Front & Europe & Industry &  2023 & \xmark & \xmark\\\midrule
         \textbf{\gradientRGB{Open MARS Dataset}{75,0,130}{193,68,14}~(Ours)}   & C\&L   & Surround & U.S. & Industry &  2024 & \cmark & \cmark \\
        \bottomrule
    \end{tabular}
    \vspace{-2mm}
\end{table*}

\section{Introduction}
\hypersetup{urlcolor=urlcolor1}
\label{sec:intro}
Autonomous driving, which has the potential to fundamentally enhance road safety and traffic efficiency, has witnessed significant advancements through AI technologies in recent years. Large-scale, high-quality, real-world data is crucial for AI-powered autonomous vehicles (AVs) to enhance their perception and planning capabilities~\cite{geiger2012we, caesar2021nuplan}: AVs can not only learn to detect objects from annotated datasets~\cite{luo2018fast} but also create safety-critical scenarios by generating digital twins based on past driving recordings~\cite{yang2023unisim}.

The pioneering KITTI dataset~\cite{geiger2012we} established the initial benchmark for tasks such as detection and tracking. Since its introduction, a number of datasets have been proposed to promote the development of self-driving; see~\cref{tab:datasets}. Two representative datasets are nuScenes~\cite{caesar2020nuscenes} and Waymo Dataset~\cite{sun2020scalability} which introduce multimodal data collected from cameras and range sensors, covering a 360-degree field of view for panoramic scene understanding. These datasets have shifted the focus from KITTI's monocular cameras, receiving wide attention in the fields of vision and robotics.

Existing driving datasets generally focus on geographical and traffic diversity without considering two practical dimensions: multiagent (\textit{collaborative}) and multitraversal (\textit{retrospective}). The \textit{collaborative} dimension highlights the synergy between multiple vehicles located in the same spatial region, facilitating their cooperative perception, prediction, and planning. The \textit{retrospective} dimension enables vehicles to enhance their 3D scene understanding by drawing upon visual memories from prior visits to the same place. Embracing these dimensions can address challenges like limited sensing capability for online perception and sparse views for offline reconstruction. Nevertheless, existing datasets are typically collected by an individual vehicle during a one-time traversal of a specific geographical location. To advance autonomous vehicle research, especially in the \textit{collaborative} and \textit{retrospective} dimensions, the research community needs a more comprehensive dataset in real-world driving scenarios. To fill the gap, we introduce the Open \textbf{MARS} Dataset, which provides \textbf{M}ulti\textbf{A}gent, multitrave\textbf{RS}al, and multimodal recordings, as shown in~\cref{fig:teaser}. \textit{All the recordings are obtained from May Mobility\footnote{\url{https://maymobility.com/}}'s autonomous vehicles operating in Ann Arbor, Michigan.}
\begin{itemize}
    \item \textbf{Multiagent.} We deploy a fleet of autonomous vehicles to navigate a designated geographical area. These vehicles can be in the same locations at the same time, allowing for collaborative 3D perception through vehicle-to-vehicle communication.
    \item \textbf{Multitraversal.} We capture multiple traversals within the same spatial area under varying lighting, weather, and traffic conditions. Each traversal may follow a unique route, covering different driving directions or lanes, resulting in multiple trajectories that provide diverse visual observations of the 3D scene.
    \item \textbf{Multimodal.} We equip the autonomous vehicle with RGB cameras and LiDAR, both with a full 360-degree field of view. This comprehensive sensor suite can enable multimodal and panoramic scene understanding.
\end{itemize}

We conduct quantitative and qualitative experiments in place recognition and neural reconstruction. More importantly, MARS introduces novel research challenges and opportunities for the vision and robotics community, including but not limited to \textit{multiagent collaborative perception and learning, unsupervised perception under repeated traversals, continual learning, neural reconstruction and novel view synthesis with multiple agents or multiple traversals}.

\section{Related Works}
\label{sec:relatedworks}
\noindent\textbf{Autonomous driving datasets.} 
High-quality datasets are crucial for advancing AI-powered autonomous driving research~\cite{caesar2020nuscenes, ettinger2021large, hu2023planning}. The seminal KITTI dataset significantly attracted research attention in robotic perception and mapping~\citep{geiger2012we,li2023voxformer, li2023sscbench,chen2023deepmapping2}. Since then, a large number of datasets have been proposed, pushing the boundaries of the field by tackling challenges in multimodal fusion, multitasking learning, adverse weather, and dense traffic~\cite{caesar2020nuscenes, huang2018apollo, liao2022kitti, pitropov2021canadian, pham20203d, xiao2021pandaset}. In recent years, researchers have proposed multiagent collaboration to get rid of the limitations in single-agent perception, \eg, frequent occlusion and long-range sparsity~\cite{li2021learning, li2022multi, li2023among, su2023uncertainty, hu2023collaboration, su2024collaborative, huang2024actformer}. Previous efforts in curating multiagent datasets are usually limited by simulated environments~\cite{xu2022opv2v, li2022v2x}. The recent V2V4Real~\citep{xu2023v2v4real} supports vehicle-to-vehicle cooperative object detection and tracking in the real world, yet the two-camera setup is insufficient for surround-view perception. Another relevant dataset, Ithaca365~\citep{diaz2022ithaca365}, provides recordings from repeated traversals of the same route in different lighting and weather conditions, yet it only uses front-view cameras for data collection. Several works collect multitraversal data for map change such as Argoverse~2 dataset~\citep{wilson2021argoverse}, and some recent works build 3D reconstruction methods or simulators based on Argoverse~2~\cite{fischer2024multi,fischer2024dynamic}. There are also several works focusing on long-term visual localization~\cite{toft2020long}, such as Oxford RobotCar Dataset~\cite{maddern20171} and CMU Seasons dataset~\cite{sattler2018benchmarking}. Yet these datasets do not consider scenarios of multiagent driving. To fill the gap, our MARS dataset provides multiagent, multitraversal, and multimodal driving recordings with a panoramic camera view; see~\cref{tab:datasets}. \textit{Notably, the continuous and dynamic operation of May Mobility's fleet of vehicles makes our MARS dataset stand out in scale and diversity, featuring hundreds of traversals at a single location and enabling collaborative driving for up to four vehicles, thereby setting a record for both traversal and agent numbers.}

\begin{table}[t]
\scriptsize
\centering
\caption{\textbf{May Mobility sensor suite specification} of each vehicle.}
    \label{tab:sensor_spec}
    \begin{tabular}{>{\raggedright\arraybackslash}p{2.5cm}>{\raggedright\arraybackslash}p{5cm}}
    \toprule
    \textbf{Sensor}&\textbf{Details}\\\midrule
         1 $\times$ LiDAR&10Hz, 128 channel, horizontal FoV 360$^\circ$, vertical FoV 40$^\circ$\\\midrule
         3 $\times$ RGB Camrea&10Hz, original resolution 1440 $\times$ 928, sampled to 720$\times$464, Horizontal FoV 60$^\circ$, Vertical FoV 40$^\circ$\\\midrule
 3 $\times$ Fisheye Camrea&10Hz, original resolution 1240 $\times$ 728, sampled to 620$\times$364, horizontal FoV 140$^\circ$, vertical FoV 88$^\circ$\\\midrule
         1 $\times$ IMU&10Hz, velocity, angular velocity, acceleration\\\midrule
         1 $\times$ GPS&10Hz, longitude, latitude, elevation\\\bottomrule
    \end{tabular}
    \vspace{-2mm}
\end{table}

\noindent\textbf{Visual place recognition.}
In the field of computer vision and robotics, visual place recognition (VPR) holds significant importance, enabling the recognition of specific places based on visual inputs~\cite{schubert2023visual}. Specifically, VPR systems function by comparing a given query data, usually an image, to an existing reference database and retrieving the most similar instances to the query. This functionality is essential for vision-based robots operating in GPS-unreliable environments. VPR techniques generally fall into two categories: traditional methods and learning-based methods. Traditional methods leverage handcrafted features~\cite{lowe1999object,BAY2008346} to generate global descriptors~\cite{Arandjelovi2013AllAV}. However, in practice, \textit{appearance variation} and \textit{limited viewpoints} can degrade VPR performance. To address the challenge of \textit{appearance variation}, learning-based methods utilize deep feature representations~\cite{Arandjelovi2015NetVLADCA,Alibey2023MixVPRFM,zhu2023r2former}. In addition to image-based VPR, video-based VPR approaches~\cite{garg2021seqnet, garg2021seqmatchnet, arcanjo2023amusic} are proposed to achieve better robustness, mitigating the \textit{limited viewpoints} with video clips. Moreover, CoVPR~\cite{li2023collaborative} introduces collaborative representation learning for VPR, bridging the gap between multiagent collaboration and place recognition, and addressing \textit{limited viewpoints} by leveraging information from collaborators. Beyond 2D image inputs, PointNetVLAD~\cite{uy2018pointnetvlad} explores point-cloud-based VPR, offering a unique perspective on place recognition. In this paper, we evaluate both single-agent VPR and collaborative VPR.

\noindent\textbf{NeRF for autonomous driving.} Neural radiance fields (NeRF)~\cite{mildenhall2021nerf} in unbounded driving scenes has recently received a lot of attention, as it not only facilitates the development of high-fidelity neural simulators~\cite{yang2023unisim} but also enables high-resolution neural reconstruction of the environment~\cite{guo2023streetsurf}. Regarding novel view synthesis (NVS), researchers have addressed the challenges such as scalable neural representations with local blocks~\cite{tancik2022block,turki2022mega}, dynamic urban scene parsing with compositional fields~\cite{ost2021neural,turki2023suds}, and panoptic scene understanding with object-aware fields~\cite{kundu2022panoptic,fu2022panoptic}. Regarding neural reconstruction, researchers have realized decent surface reconstruction based on LiDAR point cloud and image input~\cite{rematas2022urban,wang2023neural}. Meanwhile, several efforts have been made in multi-view implicit surface reconstruction without relying on LiDAR~\cite{guo2023streetsurf}. Existing methods based on NeRF are constrained by limited visual observations, often relying on sparse camera views collected along a narrow trajectory. There is significant untapped potential in leveraging additional camera perspectives, whether from multiple agents or repeated traversals, to enrich the visual input and enhance the NVS or reconstruction performance.
\vspace{-3mm}

\begin{figure}[t]
    \centering
    \includegraphics[width=\linewidth]{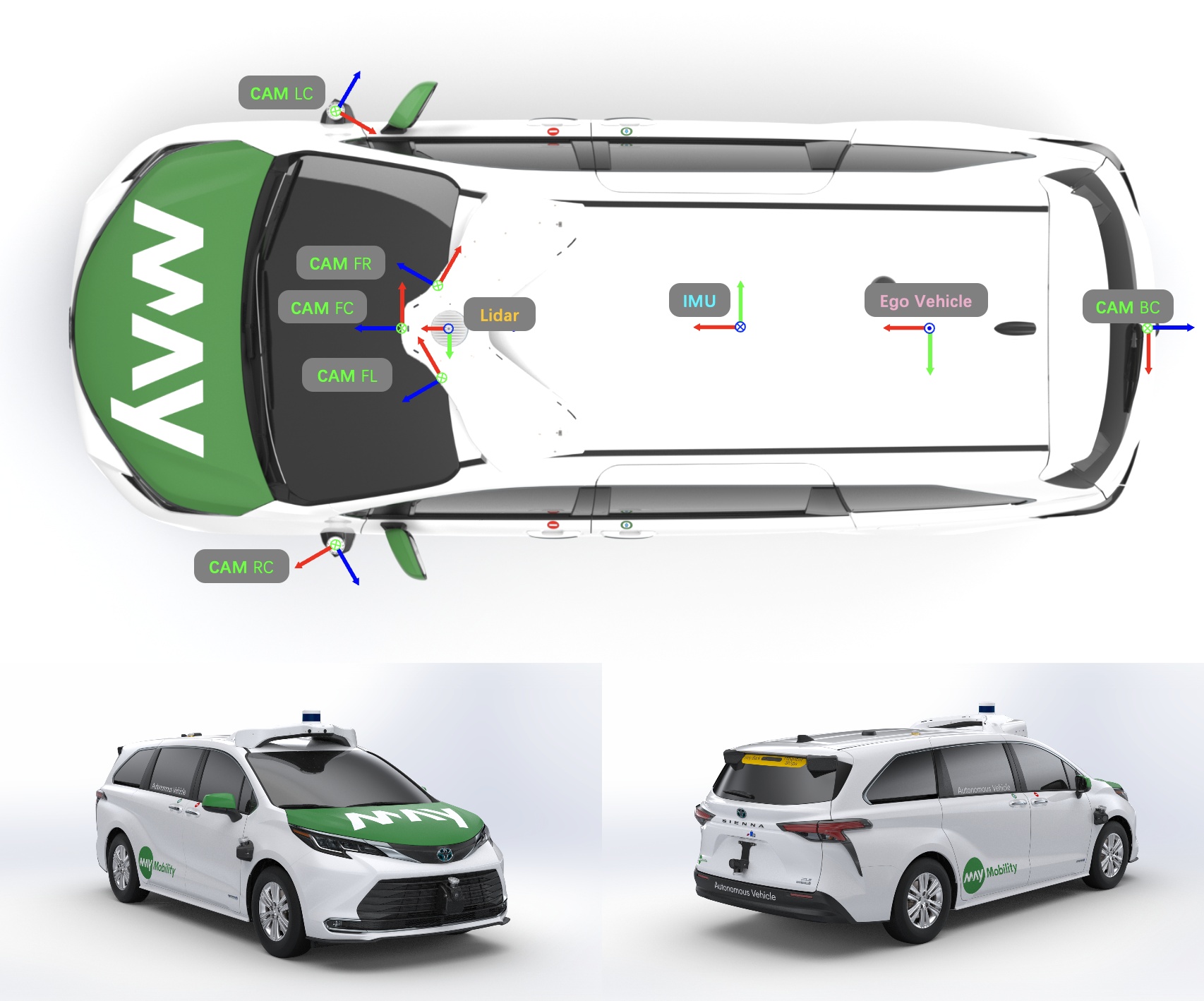}
    \caption{\textbf{Sensor setup} of the vehicle platform for data collection.}
    \label{fig:sensor_setup}
\end{figure}

\section{Dataset Curation}
\subsection{Vehicle Setup} 
\noindent\textbf{Sensor setup.} \emph{May Mobility}'s fleet of vehicles includes four Toyota Sienna, each mounted with one LiDAR, three narrow-angle RGB cameras, three wide-angle RGB fisheye cameras, one IMU, and one GPS. The sensors have various raw output frequencies, but all sensor data are eventually sampled to 10Hz for synchronization. Camera images are down-sampled to save storage. Detailed specifications of these sensors are listed in~\cref{tab:sensor_spec}.
In general, the LiDAR is located at the front top of the vehicle. The three narrow-angle cameras are located at the front, front left, and front right of the vehicle. Three fisheye cameras are on the back center, left side, and right side of the vehicle; see~\cref{fig:sensor_setup}. The IMU and GPS are located at the center top of the vehicle. The explicit extrinsic of these sensors are expressed as rotations and translations that transform sensor data from its own sensor frame to the vehicle's ego frame.  For each camera on each vehicle, we provide camera intrinsic parameters and distortion coefficients. The distortion parameters were inferred by the AprilCal calibration method~\cite{richardson2013aprilcal}.  

\noindent\textbf{Coordinate system.} There are four coordinate systems: sensor frame, ego frame, local frame, and global frame. Sensor frame represents the coordinate system whose origin is defined at the center of an individual sensor. The ego frame represents the coordinate system whose origin is defined at the center of the rear axle of an ego vehicle. The local frame represents the coordinate system whose origin is defined at the start point of an ego vehicle's trajectory of the day. The global frame is the world coordinate system.

\begin{figure}[t]
    \centering
    \includegraphics[width=0.88\linewidth]{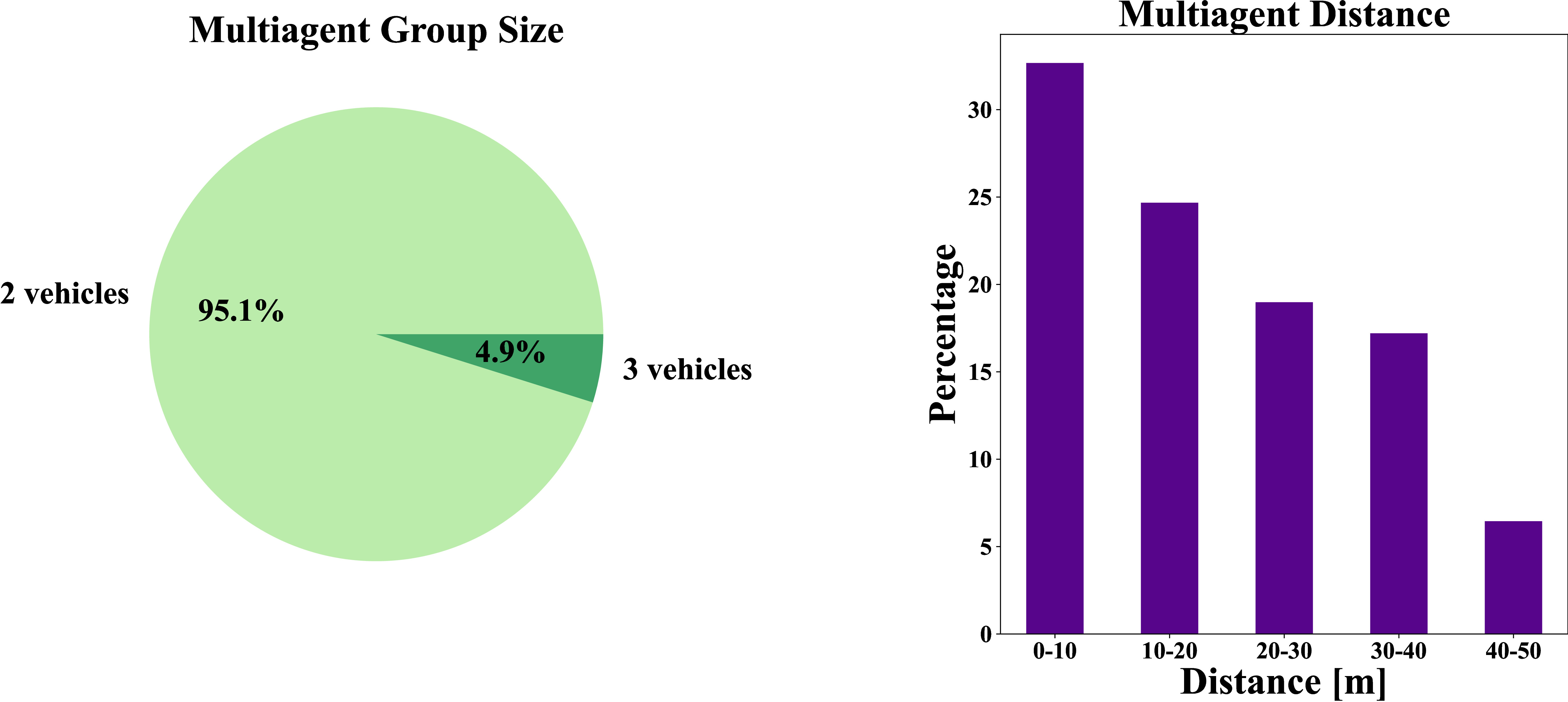}
    \vspace{-1mm}
    \caption{\textbf{Multiagent subset statistics.}}
    \label{fig:multiagent_statistics_5}
\end{figure}
\begin{figure}[t]
    \centering
    \includegraphics[width=0.85\linewidth]{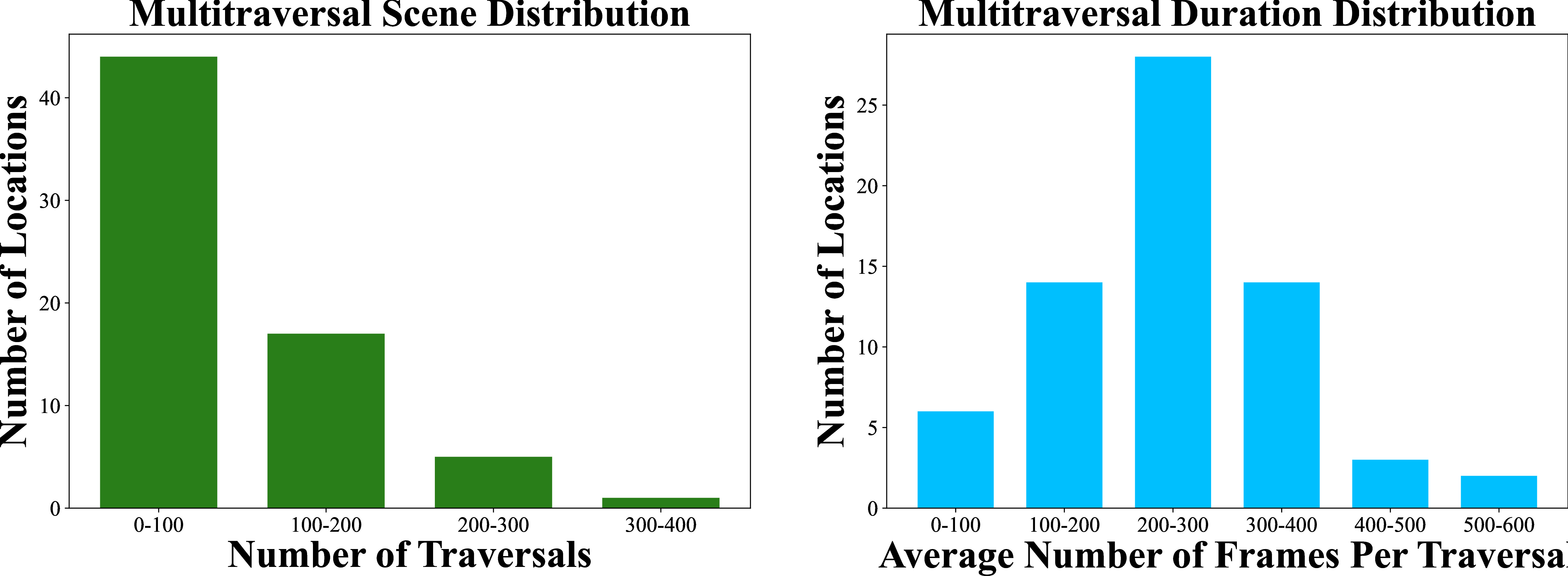}
    \caption{\textbf{Multitraversal subset statistics.}}
    \label{fig:multitraversal_statistics_1}
    \vspace{-3mm}
\end{figure}

\subsection{Data Collection}
May Mobility is currently focusing on micro-service transportation, running shuttle vehicles on fixed routes in various orders and directions. The full route is over 20 kilometers long, encompassing residential, commercial, and university campus areas with diverse surroundings in terms of traffic, vegetation, buildings, and road marks. The fleet operates every day between 2 to 8 p.m., therefore covering various lighting and weather conditions. Altogether, May Mobility's unique mode of operation enabled us to collect multitraversal and multiagent self-driving data. 

\noindent\textbf{Multitraversal data collection.}
We defined a total of 67 locations on the driving route, each spanning a circular area of a 50-meter radius. These locations cover different driving scenarios such as intersections, narrow streets, and long-straight roads with various traffic conditions. The traversals at each location take place from different directions at different times of each day, promising physically and chronologically comprehensive perceptions of the area. We determine via the vehicle's GPS location whether it is traveling through a target location, and data is collected for the full duration of the vehicle's presence within the 50-meter-radius area. Traversals are filtered such that each traversal is between 5 seconds to 100 seconds long.

\begin{figure*}
    \centering
    \includegraphics[width=0.95\linewidth]{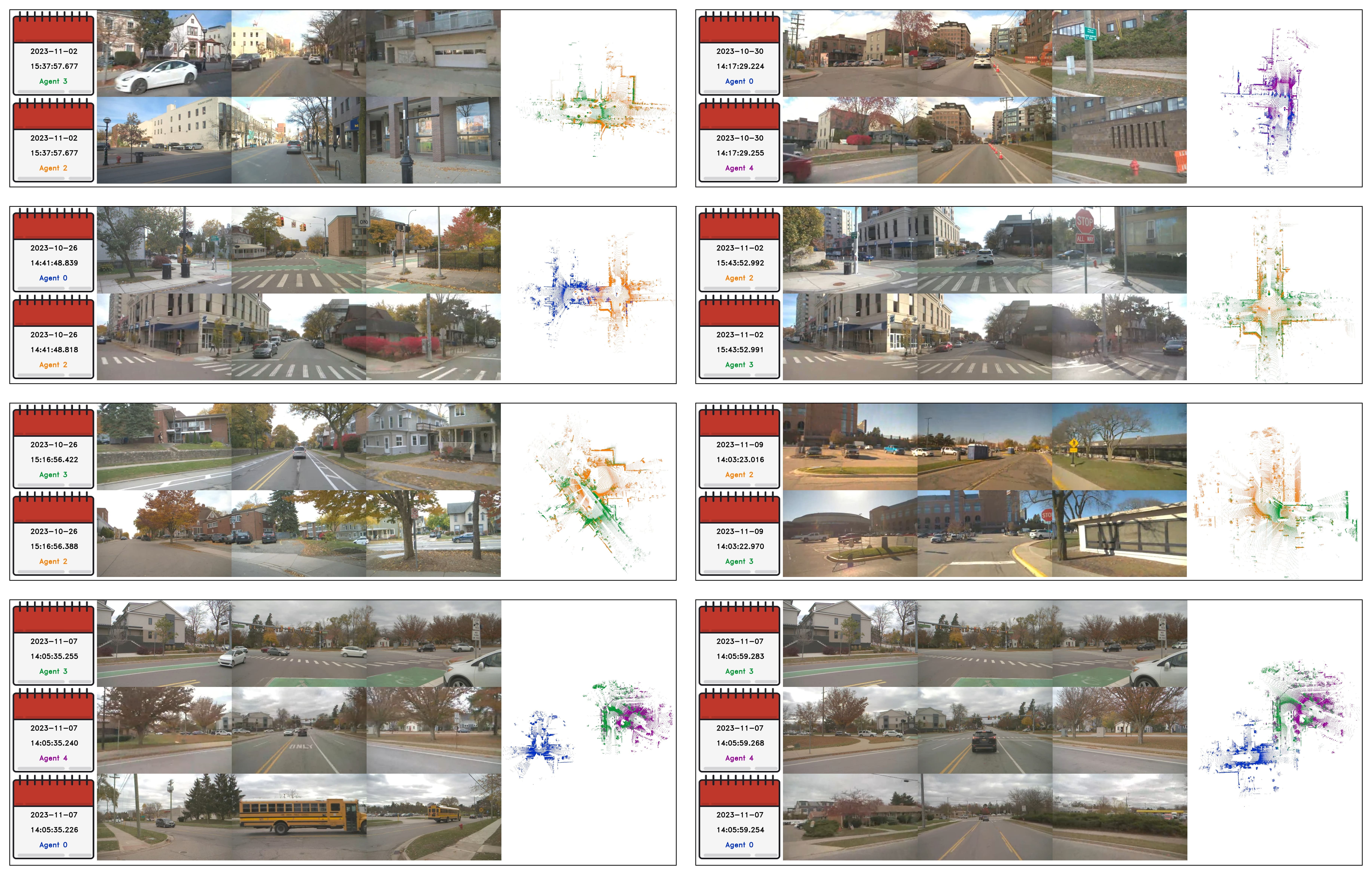}
    \caption{\textbf{Multiagent scene visualizations with three front cameras and LiDAR point clouds in bird's eye view (BEV).} Typical scenarios include straight road tailgating as well as meeting at intersections.}
    \label{fig:multiagent_fullimg}
\end{figure*}

\noindent\textbf{Multiagent data collection.}
A highlight of our dataset is that we provide real-world synchronized multi-agent collaborative perception data that delivers extremely detailed spatial coverage. Determining from vehicles' GPS coordinates, we extract 30-second-long scenes where two or more ego vehicles have been less than 50 meters away from each other for more than 9 seconds, collectively providing overlapping perceptions of the same area at the same time but from different angles. For scenes where the encountering persisted less than a full 30 seconds, the encountering segment is placed at the center of the 30-second duration, with equal amount of non-encountering time filled before and after it (\eg 20 seconds of encountering gets extended to a 30-second scene by adding 5 seconds before and 5 seconds after). Such encountering can take place anywhere around the map, constituting scenarios such as tailgating along a straight road and meeting at intersections, as shown in Fig.~\ref{fig:multiagent_fullimg}. Our method also ensures that at least one vehicle in the scene travels over 10 meters within 30 seconds.

\begin{figure}[t]
    \centering
    \includegraphics[width=0.8\linewidth]{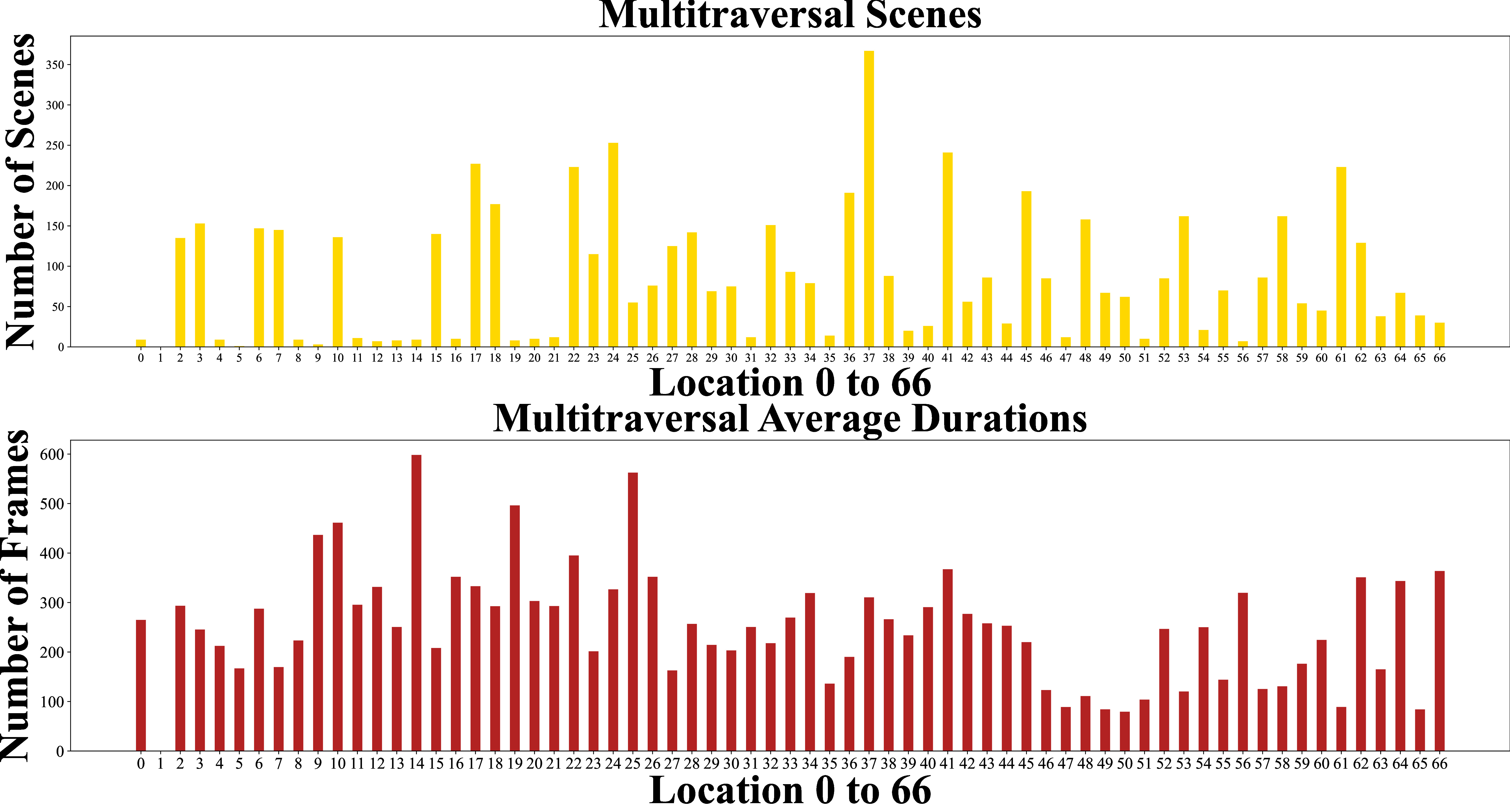}
    \vspace{-2mm}
    \caption{\textbf{Number of traversals and frames at each location.}}
    \label{fig:multitraversal_statistics_2}
    \vspace{-2mm}
\end{figure}
\begin{figure}[t]
    \centering
    \includegraphics[width=0.75\linewidth]{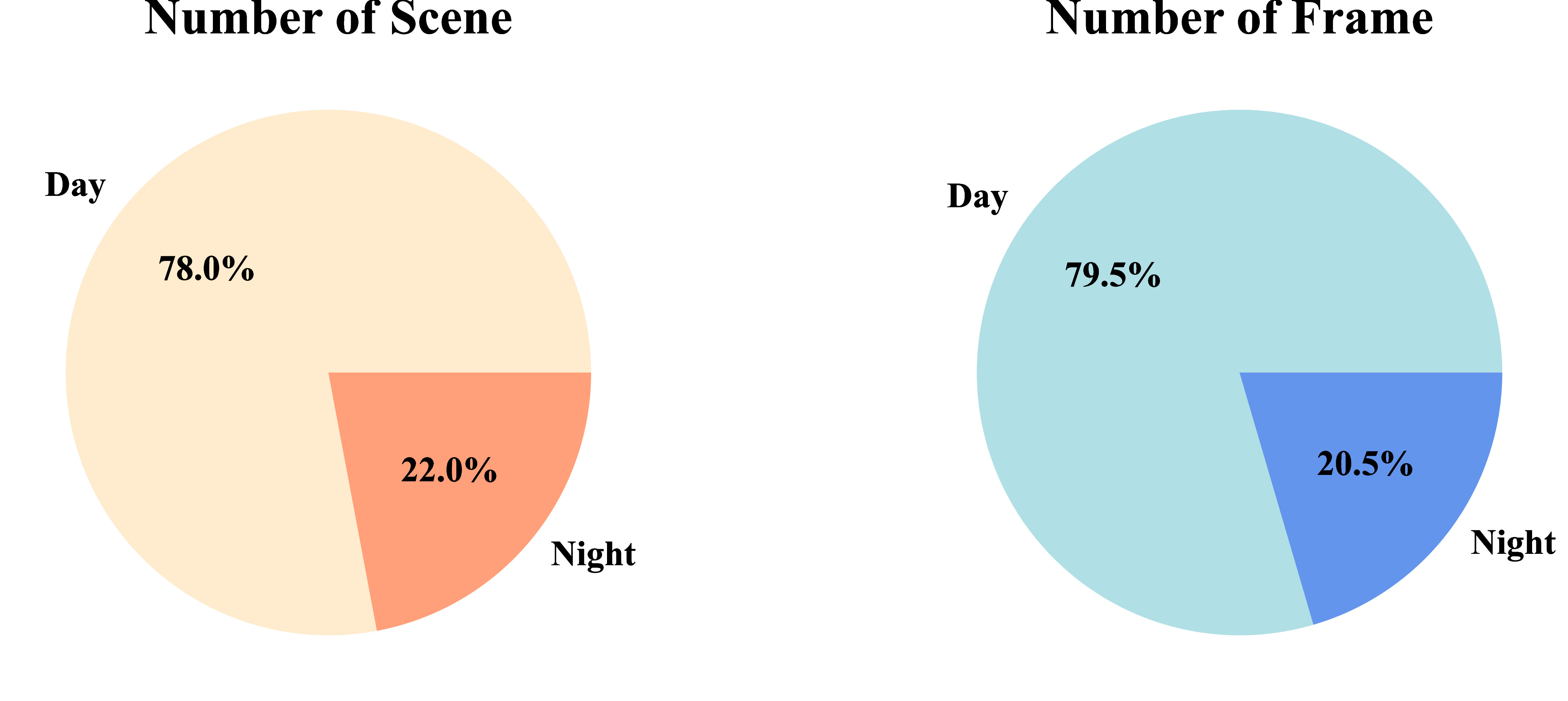}
   \vspace{-3mm}
\caption{\textbf{Ratio of day and night scenes.}}
    \label{fig:day_night}
    \vspace{-4mm}
\end{figure}

\subsection{Dataset Statistics}

The multitarversal subset covers data from 26 different days between October 4th, 2023 and March 8th, 2024, 4 of which were rainy. We collected a total of 5,757 traversals containing over 1.4 million frames of images for each camera and 360-degree LiDAR point clouds. Among the 67 locations, 48 have over 20 traversals, 23 over 100 traversals, and 6 over 200 traversals. Each traversal has 250 frames (25 seconds) on average, with the majority of traversals containing 100 to 400 frames (10 to 40 seconds). The specific distributions of traversals and frames across all locations are shown in~\cref{fig:multitraversal_statistics_1} and~\cref{fig:multitraversal_statistics_2}.
The muitlagent subset covers data from 20 different days between October 23rd, 2023 and March 8th, 2024. We collected 53 scenes of 30-second duration, stably involving 297 to 300 frames in each scene, accounting for over 15,000 frames of images and LiDAR point clouds in total. Among the 53 scenes, 52 involve two vehicles, and 1 involves three vehicles. The distance between each pair of ego vehicles is analyzed for every frame. The distribution demonstrates that encountering takes place mostly with two vehicles being less than 50 meters away from each other, as shown in~\cref{fig:multiagent_statistics_5}.


\section{Benchmark Task and Model}
\label{sec:benchmark}

\subsection{Place Recognition}

\textbf{Problem definition.} We consider a set of queries $\mathbf{Q}$ with $M$ images and a reference database $\mathbf{D}$ with $N$ images. In this task, the objective is to find $I_r \in \mathbf{D}$ given $I_q \in \mathbf{Q}$ such that $I_q$ and $I_r$ are captured at the same location.

\noindent\textbf{Evaluation metric.} We adopt recall at K as our evaluation metric for VPR. For a query image $I_q$, we select K reference images with Top-K cosine similarities between $X_q$ and $\{X_r\}_{r=1}^{N}$. If at least one of the selected images is captured within $S$ meters of $I_q$ ($S = 20$ in this paper), then we count it as correct. The recall at K is computed as the ratio between the total number of correct counts and $M$.

\noindent\textbf{Benchmark models.} We adopt NetVLAD~\cite{Arandjelovi2015NetVLADCA}, PointNetVLAD~\cite{uy2018pointnetvlad}, MixVPR~\cite{Alibey2023MixVPRFM}, GeM~\cite{Radenovic2017FineTuningCI}, Plain ViT~\cite{dosovitskiyimage}, and CoVPR~\cite{li2023collaborative} as benchmark models.

\begin{itemize}
    \item \textbf{NetVLAD} consists of a CNN-based backbone and a NetVLAD pooling layer. NetVLAD replaces the hard assignment in VLAD~\cite{Arandjelovi2013AllAV} with a learnable soft assignment, taking features extracted by backbones as input and generating a global descriptor.

    \item \textbf{MixVPR} consists of a CNN-based backbone and a feature-mixer. The output of the backbone is flattened to $C\times H'W'$, fed to the feature-mixer with row-wise and column-wise MLPs, flattened to a single vector, and $L^2$-normalized.

    \item \textbf{PointNetVLAD} consists of a backbone, a NetVLAD pooling, and an MLP. We reduced the output dimension of the backbone from 1024 to 256 and omitted the last MLP layer for efficient computation.

    \item \textbf{GeM} consists of a CNN-based backbone and a GeM pooling. The GeM pooling is defined as $\frac{1}{N}(\sum_{i=1}^NX_i^p)^{\frac{1}{p}}$, where $X_i$ is the patch feature, and we select p = 3 here.

    \item \textbf{Plain ViT}~\cite{dosovitskiyimage} consists of standard transformer encoder layers and a $L^2$ normalization over cls toekn.
    
    \item \textbf{CoVPR}~\cite{li2023collaborative} consists of a VPR model and a similarity-regularized fusion. The VPR model generates descriptors for the ego agent and collaborators, and the fusion module fuses them into a single descriptor.

\end{itemize}

\subsection{Neural Reconstruction}
\noindent\textbf{Problem definition.} Based on the number of available traversals, we divided the reconstruction task into two scenarios. The first is \textit{single-traversal (dynamic scene reconstruction)}, where the input is a sequence of images $\mathcal{I}=\{I_{1}, I_{2}, \cdots I_{k}\}$ captured as one traversal video. And the goal is to reconstruct photorealistic scene views, including moving objects. The second is \textit{multitraversal (environment reconstruction)}, where the input is a collection of image sequences $\{\mathcal{I}_1,\mathcal{I}_2,\cdots,\mathcal{I}_{n}:\mathcal{I}_{m}=\{I_{m,1},\cdots,I_{m,k_m}\}\}$ of the same scene. The objective in this task is to reconstruct the environment and remove dynamic objects.

\noindent\textbf{Evaluation metrics.}
Building on the methods used in earlier works~\cite{chen2023periodic}. we use PSNR, SSIM and LPIPS metrics for our experiments of dynamic reconstruction. PSNR, defined as \(PSNR = 10 \cdot \log_{10}\left(\frac{MAX_I^2}{MSE}\right)\), assesses image quality by comparing maximum pixel value \(MAX_I\) and mean squared error \(MSE\). SSIM, calculated by \(SSIM(x, y) = \frac{(2\mu_x\mu_y + c_1)(2\sigma_{xy} + c_2)}{(\mu_x^2 + \mu_y^2 + c_1)(\sigma_x^2 + \sigma_y^2 + c_2)}\), measures similarity between synthesized and ground truth images, factoring in mean, variance, and covariance. LPIPS, unlike the two metrics before, uses a pretrained neural network model to evaluate the perceptual similarity between two images. 

\noindent\textbf{Benchmark models.} For the single-traversal task, we adopt EmerNeRF~\cite{yang2023emernerf} and PVG~\cite{chen2023periodic} as benchmark models. Additionally, for comparison, we conduct experiments using iNGP~\cite{mueller2022instant} and 3DGS~\cite{kerbl3Dgaussians}, which do not directly target this problem. Regarding multitraversal reconstruction, there are no algorithms specifically designed for this task. Therefore, we adopt iNGP as the basic model. Furthermore, to enhance the model's ability to remove dynamic objects, we also test RobustNeRF~\cite{Sabour_2023_CVPR} and iNGP with Segformer~\cite{xie2021segformer}.

\begin{itemize}
    \item \textbf{Single-traversal: Dynamic scene reconstruction.}
    \begin{itemize}
        \item \textbf{EmerNeRF.} Based on neural fields, EmerNeRF is a self-supervised method for effectively learning spatial-temporal representations of dynamic driving scenes. EmerNeRF builds a hybrid world representation by breaking scenes into static and dynamic fields. By utilizing an emergent flow field, temporal information can be further aggregated, enhancing the rendering precision of dynamic components. The 2D visual foundation model features are lifted into 4D space-time to augment EmerNeRF's semantic scene understanding.
        \item \textbf{PVG.} Building upon 3DGS, PVG introduces periodic vibration into each Gaussian point to model the dynamic motion of these points. To handle the emergence and vanishing of objects, it also sets a time peak and a lifespan for each point. By learning all these parameters, along with the mean, covariance, and spherical harmonics of the Gaussians, PVG is able to reconstruct dynamic scenes in a memory-efficient way.  
    \end{itemize}
    \item \textbf{Multitraversal: Environment reconstruction.}
    \begin{itemize}
        \item \textbf{RobustNeRF} RobustNeRF replaces the loss function of the original NeRF to ignore distractors, and we consider dynamic objects as distractors in our case. Additionally, RobustNeRF applies a box kernel in its loss estimator to prevent high-frequency details from being recognized as outliers.
        \item \textbf{SegNeRF.} SegNeRF utilizes the pretrained semantic model SegFormer~\cite{xie2021segformer} to remove movable objects.
    \end{itemize}
\end{itemize}

\section{Experimental Results}
\subsection{Visual Place Recognition} 

  \begin{figure*}[t]
     \centering
     \includegraphics[width=\linewidth]{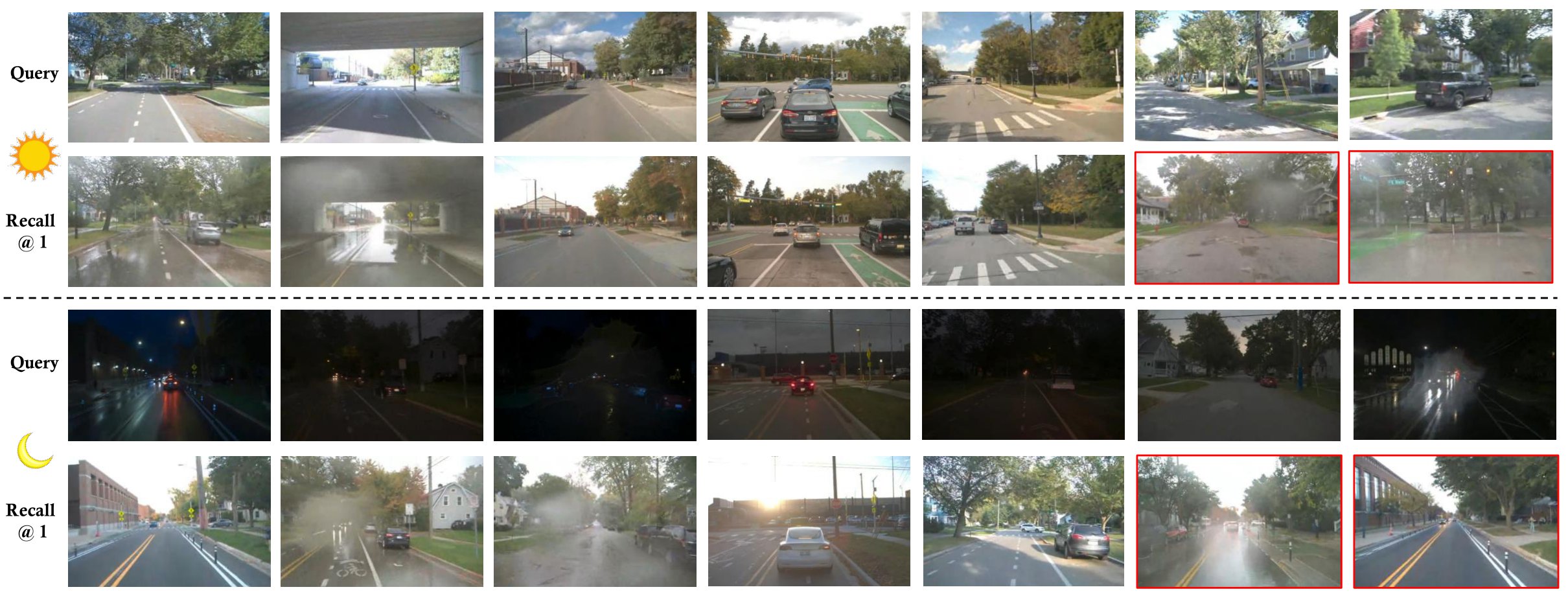}
     \caption{\textbf{Qualitative result of VPR.} We use MixVPR to obtain this qualitative result and mark incorrect results with \textcolor{red}{red} frames. Our dataset contains hard cases such as nighttime, back-lighting, and blurred cameras due to weather conditions.}
     \label{fig:vpr}
 \end{figure*}
 
\textbf{Dataset details.} We conduct experiments in VPR tasks with both multitraversal and multiagent data. In the multitraversal case, intersections numbered higher than or equal to 52 are used for testing. In the multiagent setting, scenes numbered higher than or equal to 50 are used for testing. Input images are resized to $400\times224$, and input point clouds are downsampled to 1024 points.

\noindent\textbf{Implementation details.} We evaluate our dataset on models mentioned in~\cref{sec:benchmark}, where CoVPR~\cite{li2023collaborative} is evaluated with multiagent data, and all others are evaluated with multitraversal data. Backbones are pre-trained on ImageNet1K~\cite{deng2009imagenet}. We use ResNet18~\cite{he2016deep} as the backbone for NetVLAD and CoVPR, ResNet50~\cite{he2016deep} for MixVPR and GeM, and PointNet~\cite{qi2017pointnet} for PointNetVLAD. The number of clusters in NetVLAD-based methods is 32. Models are trained with Adam~\cite{Kingma2014AdamAM} optimizer with 1e-3 lr for PointNetVLAD, 1e-4 lr for others, and 1e-4 decay rate until convergence. The batch size is 20 for NetVLAD-based methods and 10 for others. 

\noindent\textbf{Result discussions.} Quantitative results are shown in~\cref{tab:VPR}. Although GeM achieves lightweight characteristics in its pooling methods, it underperforms compared to NetVLAD with a smaller backbone. ViT demonstrates weaker performance in VPR without task-specific pooling methods, despite being a stronger backbone than ResNet. MixVPR achieves the best performance, as its feature-mixing mechanism provides richer features. PointNetVLAD, leveraging point clouds, attains better performance with smaller input sizes than NetVLAD. In the context of multiagent data, CoVPR consistently outperforms its single-agent counterparts. Qualitative results are depicted in~\cref{fig:vpr}. Our dataset encompasses both daytime and nighttime scenes, under various weather conditions such as sunny, cloudy, and rainy. Hard examples stem from nighttime scenarios and cameras affected by rain or backlighting.

\subsection{Neural Reconstruction}
\textbf{Dataset details.}
In our single-traversal dynamic scene reconstruction experiments, we selected 10 different locations, each with one traversal, aiming to capture and represent complex urban environments. For our multitraversal environment reconstruction experiments, we selected a total of 50 traversals. This comprised 10 unique locations, with 5 traversals for each location, enabling us to capture variations in illuminating conditions and weather.

 \begin{table}[t]
    \begin{minipage}{\columnwidth} 
    \centering
    \scriptsize
    \renewcommand\tabcolsep{2pt}
    \caption{\textbf{Quantitative results of VPR.}}
    \label{tab:VPR} 
    \begin{tabular}{ccccc}
        \toprule
        Data& Model  & Recall @1 & Recall @5 &  Recall @10 \\
        \midrule
         \multirow{5}{*}{Multitraversal} & NetVLAD~\cite{Arandjelovi2015NetVLADCA}  & 63.51 & 69.60 & 72.42 \\
          & MixVPR~\cite{Alibey2023MixVPRFM} & \textbf{71.73} & \textbf{75.38} & \textbf{77.20} \\
          & GeM~\cite{Radenovic2017FineTuningCI} & 61.00 & 68.47 & 71.73 \\
          & ViT~\cite{dosovitskiyimage} & 53.33 & 58.79 & 62.37 \\
          & PointNetVLAD~\cite{uy2018pointnetvlad}  & 66.45 & 72.82 & 75.91 \\\midrule
         \multirow{2}{*}{Multiagent} & NetVLAD~\cite{Arandjelovi2015NetVLADCA}  & 91.85 & 94.89 & 95.44  \\
          & CoVPR~\cite{li2023collaborative}  & \textbf{92.27} & \textbf{95.30} & \textbf{95.86}  \\
        \bottomrule 
        
    \end{tabular}
    \end{minipage}
\end{table}

 \begin{table}[t]
    \begin{minipage}{\linewidth}
    \centering
    \renewcommand\tabcolsep{3pt}
    \scriptsize
    \caption{\textbf{Quantitative results of neural reconstruction.} We compute the average PSNR, SSIM and LPIPS of ten locations to assess the reconstructed appearance. 
    }
    \label{tab:NVS} 
    \begin{tabular}{ccccc}
        \toprule
        Task &Model & PSNR $\uparrow$  & SSIM $\uparrow$ & LPIPS $\downarrow$ \\
        \midrule
        {Single-traversal}  
         & iNGP~\cite{mueller2022instant} & 28.66& 0.821  & 0.256    \\
         & 3DGS~\cite{kerbl3Dgaussians} & 27.77& 0.867  & 0.235    \\
         & EmerNeRF~\cite{yang2023emernerf} & \textbf{29.63}& 0.839  & 0.237    \\
         & PVG~\cite{chen2023periodic} & 29.28& \textbf{0.900}  & \textbf{0.197}    \\
        \midrule
        {Multitraversal}  
        & iNGP~\cite{mueller2022instant} & \textbf{26.04} & \textbf{0.759}  & \textbf{0.346}  \\
        & RobustNeRF~\cite{Sabour_2023_CVPR} & 16.17 & 0.674 & 0.459 \\
        & SegNeRF~\cite{xie2021segformer} & 24.44 & 0.748 & 0.358 \\
        \midrule
    \end{tabular}
    \end{minipage}
\end{table}

\noindent\textbf{Implementation Details.} Throughout all reconstruction experiments, we utilize 100 images from the three front cameras, along with LiDAR data, as input for each traversal. \textit{Single-traversal experiments: } Both iNGP and EmerNeRF models undergo training for 10,000 iterations utilizing the Adam~\cite{Kingma2014AdamAM} optimizer with a learning rate of 0.01 and a weight decay rate of 0.00001. For EmerNeRF, we leverage the dino feature from the DINOv2 ViT-B/14~\cite{oquab2023dinov2} foundation model. The estimator employed in this model is PropNet, incorporating linear disparity and uniform sampling. For 3DGS and PVG, we set the training iteration number to be 20000, with the learning rate the same as in the original work~\cite{chen2023periodic}. We treat 3DGS as a special case of the PVG method, with a $0$ periodic motion amplitude and an infinite lifespan, which we set to $10^6$ in our experiments.
\textit{Multitraversal experiments:} Our NeRF model in this experiment is iNGP~\cite{mueller2022instant} with image embedding and DINO features. For RobustNeRF, we implement the robust loss and patch sample as described in the original paper~\cite{Sabour_2023_CVPR}. In SegNeRF, we apply the SegFormer-B5~\cite{xie2021segformer} model, trained on the Cityscapes~\cite{Cordts2016Cityscapes} dataset. Among the 19 categories in the SegFormer model, we identify 'person', 'rider', 'car', 'truck', 'bus', 'train', 'motorcycle' and 'bicycle' as dynamic classes and generate masks for them. 

\begin{figure*}[t]
    \centering
    \includegraphics[width=0.49\linewidth]{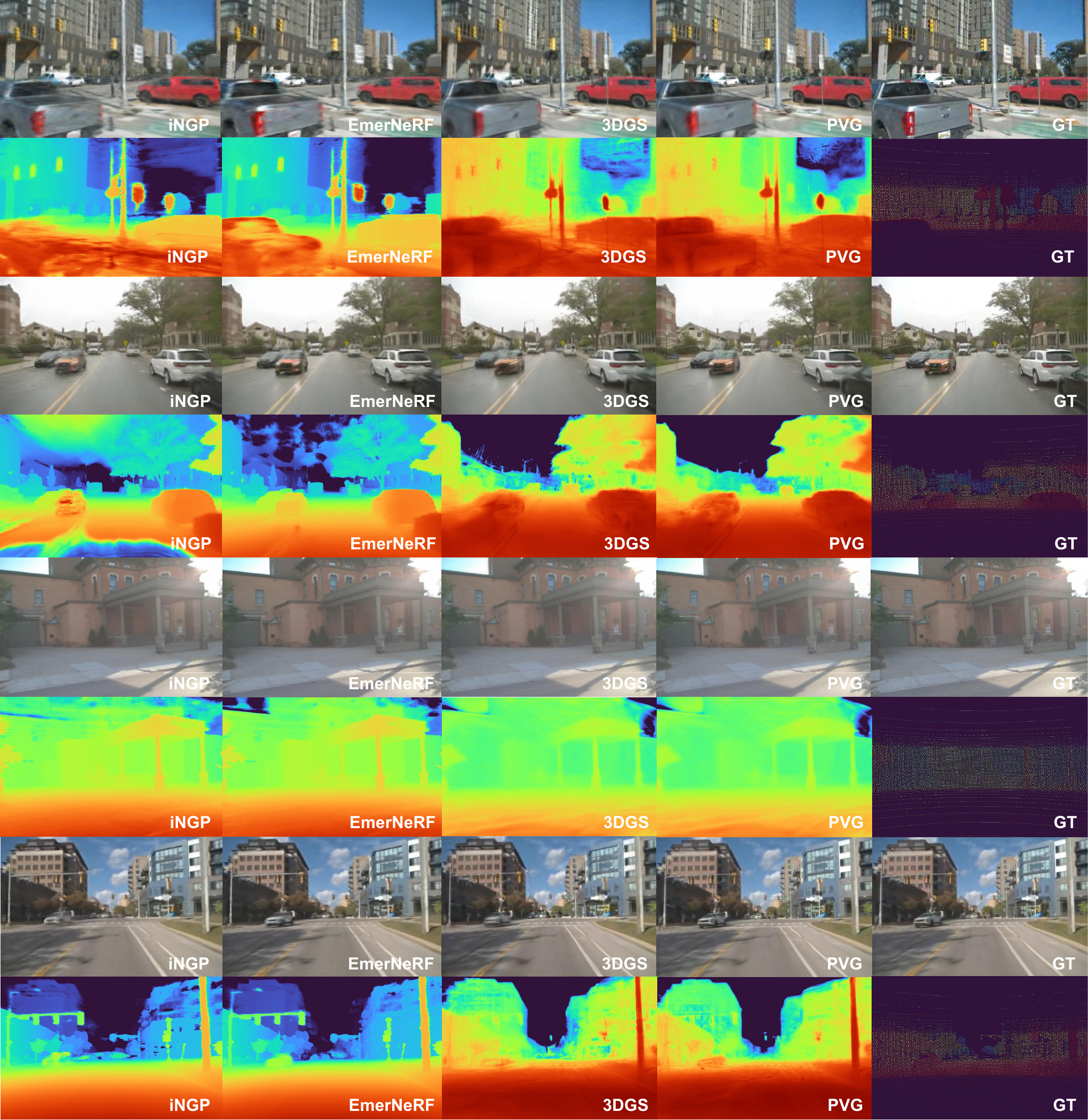}
    \includegraphics[width=0.49\linewidth]{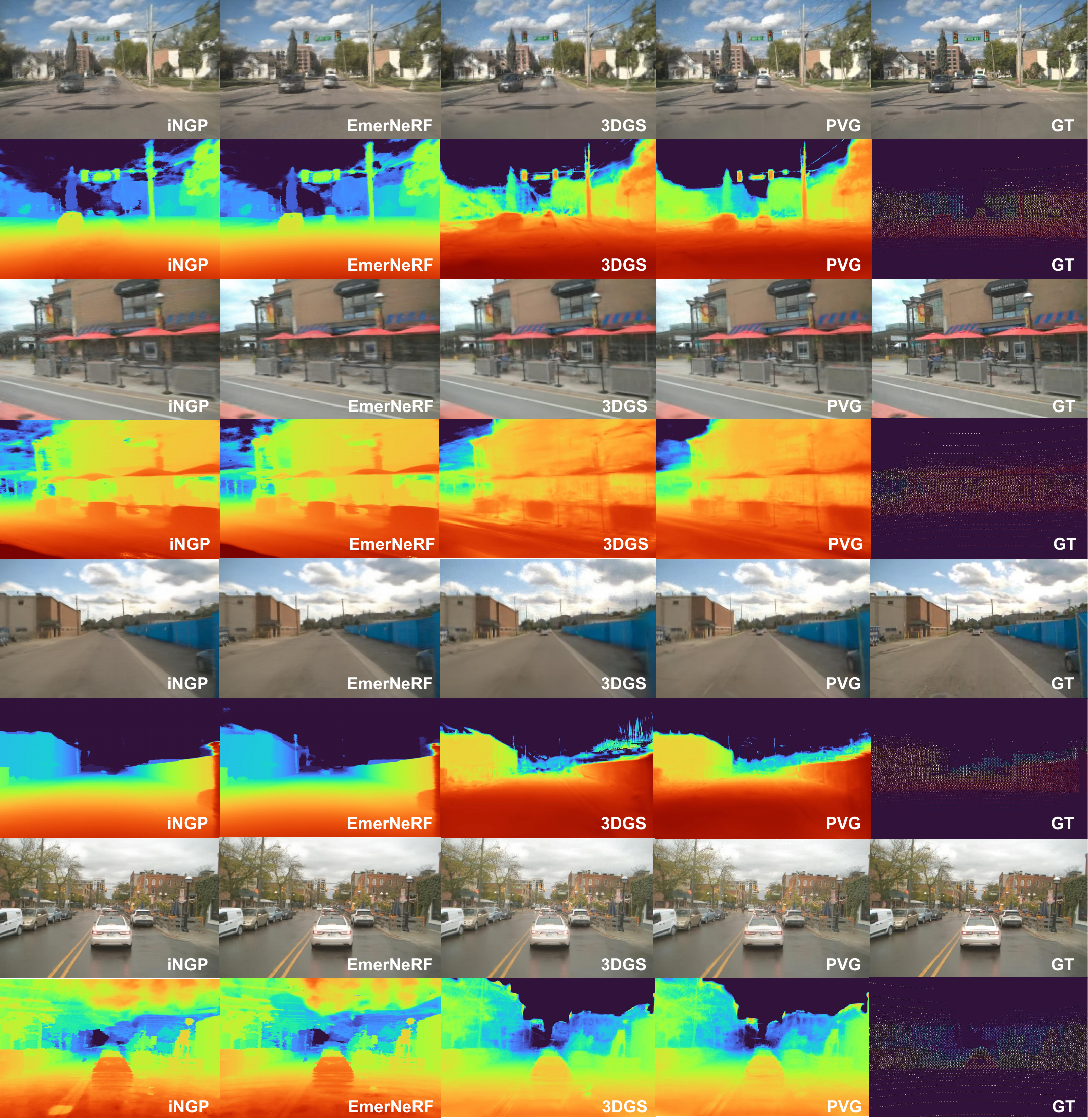}
    \caption{\textbf{Qualitative results of single-traversal reconstruction.} We stack the rendered image and the corresponding rendered depth vertically. Each column corresponds to one baseline method and the last column is the ground truth. The ground truth depth is obtained by projecting LiDAR points in the camera view.}
    \label{fig:single_traversal_reconstruction}
\end{figure*}

\noindent\textbf{Result discussions.} %
\textit{Single-traversal experiments:} Based on the results presented in~\cref{tab:NVS}, PVG achieves higher SSIM scores and better LPIPS scores, indicating enhanced structural details. This superior performance by PVG is likely attributed to its flexible Gaussian points setup, which adeptly captures linear motions, and the emergence and disappearance of objects. EmerNeRF, on the other hand, excels in PSNR. This is likely due to its novel approach of dynamic-static decomposition. As shown in~\cref{fig:single_traversal_reconstruction}, EmerNeRF and PVG both demonstrate the ability to perfectly render dynamic objects like moving cars, whereas iNGP and 3DGS exhibit relatively poor performance in this regard. \textit{Multitraversal experiments:} Thanks to image embedding, iNGP can render diversely illuminated scenes. However, it struggles with rendering dynamic objects accurately or removing them. As shown in~\cref{tab:NVS}, iNGP achieves the best similarity metrics since it preserves the most information about dynamic objects. RobustNeRF performs best in eliminating dynamic objects, albeit at the cost of rendering static objects with less detail. SegFormer, leveraging semantic information, achieves superior visual results compared to the other two methods. Yet the shadows of cars are not completely removed, likely due to the inadequate recognition of shadows by semantic segmentation models.

\section{Opportunities and Challenges}
Our MARS dataset introduces novel research opportunities with multiagent driving recordings, as well as a large number of repeated traversals of the same location. We outline several promising research directions and their associated challenges, opening new avenues for future study.

\noindent\textbf{3D reconstruction.} Repeated traversals can yield numerous camera observations for a 3D scene, facilitating correspondence search and bundle adjustment in multiview reconstruction. Our dataset can be utilized to study camera-only multitraversal 3D reconstruction, which is crucial for autonomous mapping and localization. The main challenge is to handle appearance variations and dynamic objects across repeated traversals over time. For instance, one recent work, 3D Gaussian Mapping~\cite{li2024memorize}, leverages multitraversal consensus to decompose the scene into a 3D environmental map represented by Gaussian Splatting and 2D object masks, without any external supervision.

\noindent\textbf{Neural simulation.} Multiagent and multitraversal recordings are valuable for crafting neural simulators that can reconstruct and simulate scenes and sensor data. High-fidelity simulations are essential for developing perception and planning algorithms. The main challenge lies in replicating real-world dynamics and variability, such as modeling the behavior of dynamic objects, environmental conditions, and sensor anomalies, ensuring that the simulated data provides a comprehensive and realistic testbed. For instance, one recent work proposes a neural scene representation that scales to large-scale dynamic urban areas, handles heterogeneous input data collected from multiple traversals, and substantially improves rendering speeds~\cite{fischer2024dynamic}. One concurrent work proposes a multi-level neural scene graph representation that scales to thousands of images from dozens of sequences with hundreds of fast-moving objects~\cite{fischer2024multi}.

\noindent\textbf{Unsupervised perception.} Exploiting scene priors in unsupervised 3D perception offers significant value, especially in multitraversal driving scenarios where abundant data from prior visits can enhance online perception. This approach not only facilitates a deeper understanding of the environment through the accumulation of knowledge over time but also enables unsupervised perception without the need for training with manual annotations.

\section{Conclusion} 
Our MARS dataset represents a notable advancement in autonomous vehicle research, moving beyond traditional data collection methods by integrating multiagent, multitraversal, and multimodal dimensions. MARS opens new avenues for exploring 3D reconstruction and neural simulation, collaborative perception and learning, unsupervised perception with scene priors, \etc. Future works include providing annotations for online perception tasks such as semantic occupancy prediction in scenarios of multiagent and multitraversal. We strongly believe  MARS  will establish a new benchmark in AI-powered autonomous vehicle research.

\section*{Acknowledgement}
All the raw data is obtained from May Mobility. We sincerely thank Mounika Vaka, Oscar Diec, Ryan Kuhn, Shylan Ghoujeghi, Marc Javanshad, Supraja Morasa, Alessandro Norscia, Kamil Litman, John Wyman, Fiona Hua, and Dr. Edwin Olson at May Mobility for their strong support. This work is supported by NSF Grant 2238968 and in part through the NYU IT High Performance Computing resources, services, and staff expertise.

{\small
\bibliographystyle{unsrt}
\bibliography{main}
}

\end{document}